\documentclass{article}

\PassOptionsToPackage{numbers, compress}{natbib}
\usepackage[final]{nips_2017}
\usepackage[pdftex]{graphicx}

\usepackage[utf8]{inputenc} 
\usepackage[T1]{fontenc}    
\usepackage{hyperref}       
\usepackage{url}            
\usepackage{booktabs}       
\usepackage{diagbox}

\title{A simple but tough-to-beat baseline for the \\ Fake News Challenge stance detection task}

\newcommand*{\affaddr}[1]{#1}
\newcommand*{\affmark}[1][*]{\textsuperscript{#1}}
\newcommand*{\email}[1]{\texttt{#1}}

\author{Benjamin Riedel\affmark[1], Isabelle Augenstein\affmark[1]\affmark[2], Georgios P. Spithourakis\affmark[1], Sebastian Riedel\affmark[1]\\
\affaddr{\affmark[1]Department of Computer Science, University College London, United Kingdom}\\
\affaddr{\affmark[2]Department of Computer Science, University of Copenhagen, Denmark}\\
\email{benjamin.riedel.09@ucl.ac.uk}, \\ \email{\{i.augenstein|g.spithourakis|s.riedel\}@cs.ucl.ac.uk}
}

\begin{document}

\maketitle

\begin{abstract}
    Identifying public misinformation is a complicated and challenging task. An important part of checking the veracity of a specific claim is to evaluate the stance different news sources take towards the assertion. Automatic stance evaluation, i.e. stance detection, would arguably facilitate the process of fact checking. In this paper, we present our stance detection system which claimed third place in Stage 1 of the Fake News Challenge. Despite our straightforward approach, our system performs at a competitive level with the complex ensembles of the top two winning teams. We therefore propose our system as the `simple but tough-to-beat baseline' for the Fake News Challenge stance detection task.
\end{abstract}

\section{Introduction}

Automating stance evaluation has been suggested as a valuable first step towards assisting human fact checkers to detect inaccurate claims. The Fake News Challenge initiative thus recently organised the first stage of a competition (FNC-1) to foster the development of systems for automatically evaluating what a news source is saying about a particular issue~\cite{fnc}.

\begin{figure}[h]
  \centering
  \includegraphics[width=1.0\textwidth]{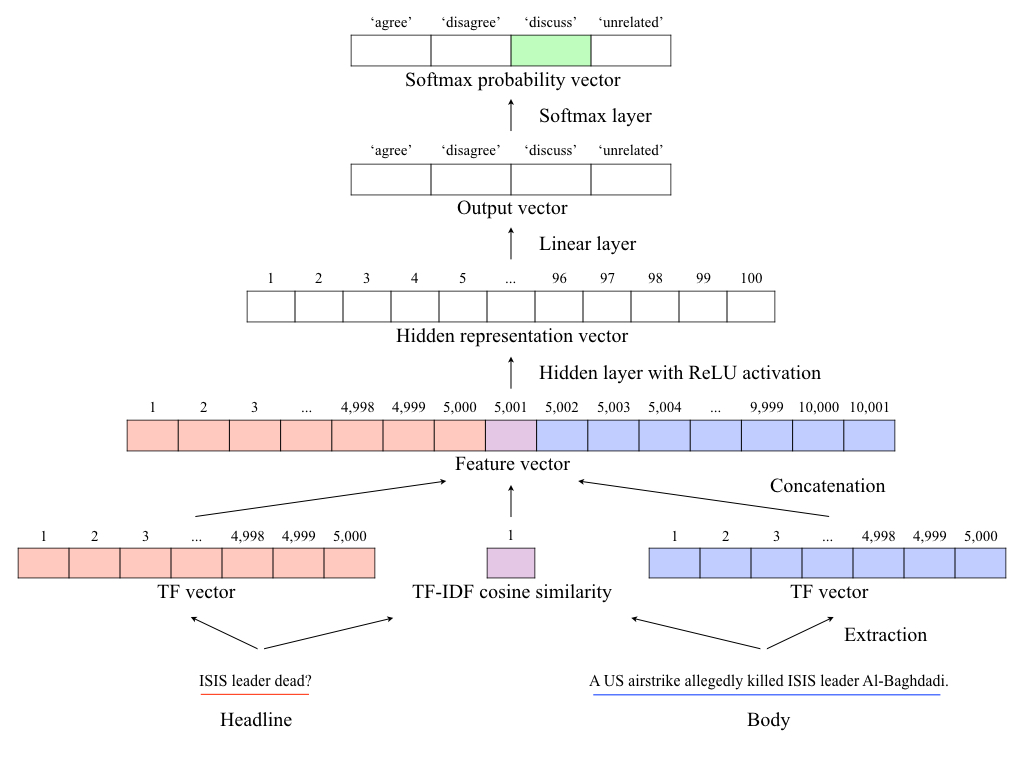}
  \caption{Schematic diagram of UCLMR's system.}
  \label{fig: model}
\end{figure}

More specifically, FNC-1 involved developing a system that, given a news article headline and a news article body, estimates the stance of the body towards the headline. The stance label to be assigned could be one of the set: `agree', `disagree', `discuss', or `unrelated' (see example of Figure~\ref{fig: model}).
More information on the FNC-1 task, rules, data, and evaluation metrics can be found on the official website: \hyperlink{http://www.fakenewschallenge.org}{\texttt{fakenewschallenge.org}}.

The goal of this short paper is to present a description of UCL Machine Reading's (UCLMR) system employed during FNC-1, a summary of the system's performance, a brief overview of the competition, and our work going forward.

\section{System description}

The single, end-to-end stance detection system consists of lexical and similarity features passed through a multi-layer perceptron (MLP) with one hidden layer.
Although relatively simple in nature, the system performs on par with more elaborate, ensemble-based systems of other teams \cite{talos, athene} (see Section~\ref{comp}).

The code for our system and instructions on how to reproduce our submission are available at UCLMR's public GitHub repository: \hyperlink{http://www.github.com/uclmr/fakenewschallenge}{\texttt{github.com/uclmr/fakenewschallenge}}.

\subsection{Representations and features}

We use two simple bag-of-words (BOW) representations for the text inputs: term frequency (TF) and term frequency-inverse document frequency (TF-IDF)~\cite{tf, tfidf}.
The representations and feature extracted from the headline and body pairs consist of only the following:

\begin{itemize}
  \item The TF vector of the headline;
  \item The TF vector of the body;
  \item The cosine similarity between the $\ell_{2}$-normalised TF-IDF vectors of the headline and body.
\end{itemize}

We tokenise the headline and body texts as well as derive the relevant vectors using \texttt{scikit-learn}~\cite{scikit}.

Different vocabularies are used for calculating the TF and TF-IDF vectors. For the TF vectors, we extract a vocabulary of the 5,000 most frequent words in the training set and exclude stop words (the \texttt{scikit-learn} stop words for the English language with negation terms removed). For the TF-IDF vectors, a vocabulary of the 5,000 most frequent words is defined on both the training and test sets and the same set of stop words is excluded.

The TF vectors and the TF-IDF cosine similarity are concatenated in a feature vector of total size 10,001 and fed into the classifier.

\bigskip
\bigskip

\subsection{Classifier}

The classifier is a MLP~\cite{mlp} with one hidden layer of 100 units and a softmax on the output of the final linear layer. We use the rectified linear unit (ReLU) activation function~\cite{relu} as non-linearity for the hidden layer. The system predicts with the highest scoring label (`agree', `disagree', `discuss', or `unrelated'). The classifier as described is fully implemented in \texttt{TensorFlow}~\cite{tensorflow}.

\subsection{Training}

Our training objective was to minimise the cross entropy between the system's softmax probabilities and the true labels. For regularisation of the system, we added $\ell_{2}$ regularisation of the MLP weights to the objective and applied dropout~\cite{drop} on the output of both perceptron layers during training.

We trained in mini-batches over the entire training set with back-propagation using the Adam optimiser~\cite{adam} and gradient clipping by a global norm clip ratio~\cite{grad}. All of the aforementioned were implemented in \texttt{TensorFlow}~\cite{tensorflow}.

Training was stopped early based on a qualitative criterion with respect to the plateau of the loss on the training set and the mean performance of the system on 50 random splits of the data into training and hold-out sets as defined in the official baseline setup~\cite{fncbase}.

\subsection{Hyperparameters}

The full set of hyperparameters of the system, their labels, their descriptions, the ranges of values considered during tuning, and corresponding optimised values are provided in Table~\ref{tab: hyperparam}.
The hyperparameters were tuned during development using random search on a grid of combinations of values considered and cross-validation on various splits of the data.

\begin{table}[h]
  \caption{Details on hyperparameters of UCLMR's system.}
  \label{tab: hyperparam}
  \centering
  \begin{tabular}{llll}
    \toprule
    Label & Description & Range & Optimised \\
     \midrule
     \texttt{lim\_unigram} & BOW vocabulary size & 1,000 - 10,000 & 5,000 \\ 
     \texttt{hidden\_size} & MLP hidden layer size & 50 - 600 & 100 \\  
     \texttt{train\_keep\_prob} & 1 - dropout on layer outputs & 0.5 - 1.0 & 0.6 \\
     \texttt{l2\_alpha} & $\ell_{2}$ regularisation strength & 0.1 - 0.0000001 & 0.0001 \\
     \texttt{learn\_rate} & Adam learning rate & 0.1 - 0.001 & 0.01 \\
     \texttt{clip\_ratio} & Global norm clip ratio & 1 - 10 & 5 \\
     \texttt{batch\_size} & Mini-batch size & 250 - 1,000 & 500 \\
     \texttt{epochs} & Number of epochs & $\leq1,000$ & 90 \\
    \bottomrule
  \end{tabular}
\end{table}

\section{Results}
\label{results}

Submissions to the competition were evaluated with respect to the FNC-1 score, as defined in the official evaluation metrics~\cite{fnc}. Our submission achieved a FNC-1 score of 81.72\%. 

The performance of our system is summarised by below confusion matrix for the labels of and predictions on the final test set (see Table \ref{tab: results}). We conclude that although our system performs satisfactorily in general, this can mainly be attributed to the close to perfect classification of the instances into `related' and `unrelated' headline/body pairs (accuracy: 96.55\%) and the more or less default `discuss' classification of the `related' instances.

\begin{table}[ht]
\caption{Confusion matrix of UCLMR's FNC-1 submission.}
\label{tab: results}
\centering
\begin{tabular}{lcccccc}
\toprule
\diagbox{True}{Pred.} & `agree' & `disagree' & `discuss' & `unrelated' & Overall & \% Accuracy \\
\midrule
`agree' & 838 & 12 & 939 & 114 & 1,903 & 44.04 \\
`disagree' & 179 & 46 & 356 & 116 & 697 & 6.60 \\
`discuss' & 523 & 46 & 3,633 & 262 & 4,464 & 81.38 \\
`unrelated' & 53 & 3 & 330 & 17,963 & 18,349 & 97.90 \\
\midrule
Overall & 1,593 & 107 & 5,258 & 18,455 & 25,413 & 88.46 \\
\bottomrule
\end{tabular}
\end{table}

Our system's performance with respect to the `agree' label is average at best, whereas the system's accuracy on the `disagree' test examples is clearly quite poor. The disappointing performance is noteworthy since these two labels are arguably the most interesting in the FNC-1 task and the most relevant to the superordinate goal of automating the stance evaluation process.

\section{Competition}
\label{comp}

A total of 50 teams actively participated in FNC-1. The final top 10 leader board (see Table~\ref{tab: leaderboard}) shows our submission (UCLMR) placed in third position. The official baseline achieved a FNC-1 score of 75.20\% on the test data set~\cite{fncbase} and is included in Table~\ref{tab: leaderboard} for reference.

\begin{table}[h]
\caption{Top 10 FNC-1 leader board. UCLMR submission in \textbf{bold}.}
\label{tab: leaderboard}
\centering
\begin{tabular}{lc}
\toprule
Team & \% FNC-1 score \\
\midrule
SOLAT in the SWEN & 82.02 \\
Athene & 81.97 \\
\textbf{UCL Machine Reading} &  \textbf{81.72} \\
Chips Ahoy! & 80.21 \\
CLUlings & 79.73 \\
unconscious bias & 79.69 \\
OSU & 79.65 \\
MITBusters & 79.58 \\
DFKI LT & 79.56 \\
GTRI - ICL & 79.33 \\
\midrule
Official baseline & 75.20 \\
\bottomrule
\end{tabular}
\end{table}

The competition was won by team `SOLAT in the SWEN' from Talos Intelligence, a threat intelligence subsidiary of Cisco Systems, and the second place was taken by team `Athene' consisting of members from the Ubiquitous Knowledge Processing Lab and the Adaptive Preparation of Information from Heterogeneous Sources Research Training Group at Technische Universität Darmstadt (TU Darmstadt). The respective FNC-1 scores of these teams were 82.02\% and 81.97\%.

The team from Talos Intelligence employed a 50/50 weighted average ensemble of (i) two one-dimensional convolutional neural networks on respectively word embeddings of the headline and body feeding into a MLP with three hidden layers and (ii) five overarching sets of features passed into gradient boosted decision trees~\cite{talos}.

The team from TU Darmstadt used an ensemble of five separate MLPs, each with seven hidden layers and fed with seven overarching sets of features. Predictions were based on the hard vote of the five separate, randomly initialised MLPs~\cite{athene}.

The submission of our team performed almost on par with the top two teams and with considerable distance to the remaining teams as well as the official baseline. In contrast to other submissions, we achieved competitive results with a simple, single, end-to-end system.

Discussions with other teams, including those from Talos Intelligence and TU Darmstadt, revealed that the test set performance of other systems on the labels of key interest (`agree' and `disagree') was not much better, if at all.

\section{Future work}

Our goal going forward is to carry out in-depth analyses of our system.
The added benefit of our straightforward setup, as opposed to more sophisticated neural network architectures, is that it provides an opportunity to try to understand how it works, what contributes to its performance, and what its limitations are.

A particular focus of these analyses will be to try and identify what the mediocre performance of the system with respect to the `agree' and `disagree' labels can potentially be traced back to, next to the limited size of the data set overall and the small number of instances of the labels of specific interest.

Notwithstanding this, we would like to propose our system as the `simple but tough-to-beat baseline'~\cite{baseline} for the FNC-1 stance detection task given the system's competitive performance and basic implementation. We accordingly welcome researchers and practitioners alike to employ, improve, and/or extend our work thus far.

\subsubsection*{Acknowledgements}

We would like to thank Richard Davis and Chris Proctor at Stanford University for the description of their FNC-1 development efforts~\cite{stanford}. The system presented here is based on their setup.

It should also be noted that Akshay Agrawal, Delenn Chin and Kevin Chen from Stanford University were the first to propose their FNC-1 development efforts as the ‘simple but tough-to-beat baseline’~\cite{stanford2}. Our exchange with them was informative and highly appreciated.

Furthermore, we are grateful for insightful discussions with the following individuals during system development and the official competition.

\begin{itemize}
    \item Florian Mai at the Christian-Albrechts Universität zu Kiel
    \item Anna Seg from the FNC-1 team `annaseg'
    \item James Thorne at the University of Sheffield
    \item Sean Bird from Talos Intelligence
    \item Andreas Hanselowski at the Technische Universität Darmstadt
    \item Jingbo Shang at the University of Illinois at Urbana-Champaign
\end{itemize}

This work was supported by a Marie Curie Career Integration Award, an Allen Distinguished Investigator Award, and Elsevier. 

\bibliography{uclmr_fnc}

\end{document}